\documentclass[a4paper]{llncs}
\usepackage[utf8]{inputenc}
\usepackage[english]{babel}
\usepackage{graphicx}
\usepackage{subfiles}
\usepackage[font=small]{caption}
\usepackage{subcaption}
\usepackage{blindtext}
\usepackage{amsmath, amsfonts, amssymb}
\usepackage{listings}
\usepackage{url}
\usepackage{hyperref}
\usepackage{fancyhdr}

\graphicspath{{figures/}}
\usepackage{fancyhdr}

\let\oldref\ref
\renewcommand{\ref}[1]{(\oldref{#1})}
\urldef{\mailsa}\path|{kamyar.nazeri, eric.ng, mehran.ebrahimi}@uoit.ca|

\begin{document}

\title{Image Colorization using \\ Generative Adversarial Networks}
\titlerunning{ }

\author{Kamyar Nazeri, Eric Ng, and Mehran Ebrahimi}
\authorrunning{ }

\institute{Faculty of Science, University of Ontario Institute of Technology\\
	2000 Simcoe Street North, Oshawa, Ontario, Canada ~  L1H 7K4\\
	\mailsa\\
	\url{http://www.ImagingLab.ca/}
}

\maketitle

\begin{abstract}
Over the last decade, the process of automatic image colorization has been of significant interest for several application areas including restoration of aged or degraded images. This problem is highly ill-posed due to the large degrees of freedom during the assignment of color information. Many of the recent developments in automatic colorization involve images that contain a common theme or require highly processed data such as semantic maps as input. In our approach, we attempt to fully generalize the colorization procedure using a conditional Deep Convolutional Generative Adversarial Network (DCGAN). The network is trained over datasets that are publicly available such as CIFAR-10 and Places365. The results between the generative model and traditional deep neural networks are compared. 
\end{abstract}

\section{Introduction}

	The automatic colorization of grayscale images has been an active area of research in machine learning for an extensive period of time. This is due to the large variety of applications such color restoration and image colorization for animations. In this manuscript, we will explore the method of colorization using generative adversarial networks (GANs) proposed by Goodfellow et al. \cite{goodfellow2014generative}. The network is trained on the datasets CIFAR-10 and Places365 \cite{zhou2016places} and its results will be compared with those obtained using existing convolutional neural networks (CNN). \\
	Models for the colorization of grayscales began back in the early 2000s. In 2002, Welsh et al. \cite{welsh2002transferring} proposed an algorithm that colorized images through texture synthesis. Colorization was done by matching luminance and texture information between an existing color image and the grayscale image to be colorized. However, this proposed algorithm was defined as a forward problem, thus all solutions were deterministic. Levin et al. \cite{levin2004colorization} proposed an alternative formulation to the colorization problem in 2004. This formulation followed an inverse approach, where the cost function was designed by penalizing the difference between each pixel and a weighted average of its neighboring pixels. Both of these proposed methods still required significant user intervention which made the solutions less than ideal. \\
	In \cite{isola2016image}, a colorization method was proposed by comparing colorization differences between those generated by convolutional neural networks and GAN. The models in the study not only learn the mapping from input to output image, but also learn a loss function to train this mapping. Their approach was effective in ill-posed problems such as synthesizing photos from label maps, reconstructing objects from edge maps, and colorizing images. We aim to extend their approach by generalizing the colorization procedure to high resolution images and suggest training strategies that speed up the process and greatly stabilize it.

\section{Generative Adversarial Network}

\begin{figure}[!htb]
		\centering
		\includegraphics[width=7cm]{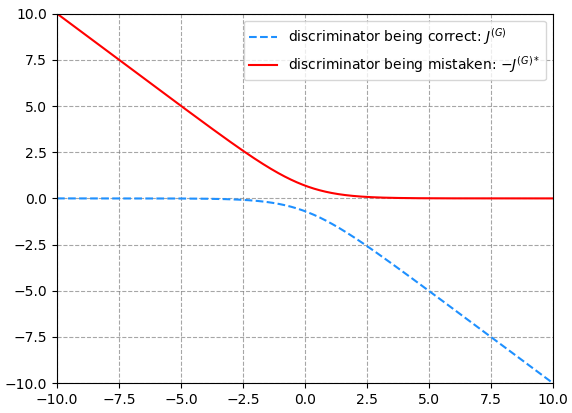}
		\caption{Comparison of cost functions $J^{(G)}$ (dashed blue) and $-J^{(G)^*}$ (red).}
		\label{fig:cost}
	\end{figure}

	In 2014, Goodfellow et al. \cite{goodfellow2014generative} proposed a new type of generative model: generative adversarial networks (GANs). A GAN is composed of two smaller networks called the generator and discriminator. As the name suggests, the generator's task is to produce results that are indistinguishable from real data. The discriminator's task is to classify whether a sample came from the generator's model distribution or the original data distribution. Both of these subnetworks are trained simultaneously until the generator is able to consistently produce results that the discriminator cannot classify. \\
	The architectures of the generator and discriminator both follow a multilayer perceptron model. Since colorization is a class of image translation problems, the generator and discriminator are both convolutional neural networks (CNNs). The generator is represented by the mapping $G(z; \theta_G)$, where $z$ is a noise variable (uniformly distributed) that acts as the input of the generator. Similarly, the discriminator is represented by the mapping $D(x; \theta_D)$ to produce a scalar between 0 and 1, where $x$ is a color image. The output of the discriminator can be interpreted as the probability of the input originating from the training data. These constructions of $G$ and $D$ enable us to determine the optimization problem for training the generator and discriminator: $G$ is trained to minimize the probability that the discriminator makes a correct prediction in generated data, while $D$ is trained to maximize the probability of assigning the correct label. Mathematically, this can be expressed as
	\begin{equation}
		\min_{\theta_G} J^{(G)} (\theta_D, \theta_G) = \min_{\theta_G} \mathbb{E}_z \left[ \log (1 - D(G(z))) \right],
		\label{eq:cost_gen}
	\end{equation}
	\begin{equation}
		\max_{\theta_D} J^{(D)} (\theta_D, \theta_G) = \max_{\theta_D} \left( \mathbb{E}_x \left[ \log (D(x)) \right] + \mathbb{E}_z \left[ \log (1 - D(G(z))) \right] \right).
	\end{equation}
	The above two equations provide the cost functions required to train a GAN. In literature, these two cost functions are often presented as a single minimax game problem with the value function $V(G, D)$:
	\begin{equation}
		\min_G \max_D V(G,D) = \mathbb{E}_x \left[ \log D(x) \right] + \mathbb{E}_z \left[ \log (1 - D(G(z))) \right].
	\end{equation}
	In our model, we have decided to use an alternate cost function for the generator.
	In equation \ref{eq:cost_gen}, the cost function is defined by minimizing the probability of the discriminator being correct. However, this approach presents two issues: 1) If the discriminator performs well during training stages, the generator will have a near-zero gradient during back-propagation. This will tremendously slow down convergence rate because the generator will continue to produce similar results during training. 2) The original cost function is a strictly decreasing function that is unbounded below. This will cause the cost function to diverge to $- \infty$ during the minimization process. \\
	To address the above issues, we have redefined the generator's cost function by maximizing the probability of the discriminator being mistaken, as opposed to minimizing the probability of the discriminator being correct. The new cost function was suggested by Goodfellow at NIPS 2016 Tutorial \cite{goodfellow2016nips} as a heuristic, non-saturating game, and is presented as:
	\begin{equation}
		\max_{\theta_G} J^{(G)^*} (\theta_D, \theta_G) = \max_{\theta_G} \mathbb{E}_z \left[ \log (D(G(z))) \right],
	\end{equation}
	which can also be written as the minimization problem:
	\begin{equation}
		\min_{\theta_G} -J^{(G)^*} (\theta_D, \theta_G) = \min_{\theta_G} - \mathbb{E}_z \left[ \log (D(G(z))) \right].
		\label{eq:cost_gen_min}
	\end{equation}
	The comparison between the cost functions in equations \ref{eq:cost_gen} and \ref{eq:cost_gen_min} can be visualized in figure \ref{fig:cost} by the blue and red curves respectively.
	\vspace{-10px}
	\vspace{-10px}
	In addition, the cost function was further modified by using the $\ell^1$-norm in the regularization term \cite{isola2016image}. This produces an effect where the generator is forced to produce results that are similar to the ground truth images. This will theoretically preserve the structure of the original images and prevent the generator from assigning arbitrary colors to pixels just to ``fool'' the discriminator. The cost function takes the form
	\begin{equation}
		\min_{\theta_G} J^{(G)^*} (\theta_D, \theta_G) = \min_{\theta_G} - \mathbb{E}_z \left[ \log (D(G(z))) \right] + \lambda \lVert G(z) - y \rVert_1
	\end{equation}
	where $\lambda$ is a regularization parameter and $y$ is the ground truth color labels. 

\subsection{Conditional GAN}

	In a traditional GAN, the input of the generator is randomly generated noise data $z$. However, this approach is not applicable to the automatic colorization problem because grayscale images serve as the inputs of our problem rather than noise. This problem was addressed by using a variant of GAN called conditional generative adversarial networks \cite{mirza2014conditional}. Since no noise is introduced, the input of the generator is treated as zero noise with the grayscale input as a prior, or mathematically speaking, $G(\mathbf{0}_z|x)$. In addition, the input of the discriminator was also modified to accommodate for the conditional network. By introducing these modifications, our final cost functions are as follows:
	\begin{equation}
		\min_{\theta_G} J^{(G)} (\theta_D, \theta_G) = \min_{\theta_G} - \mathbb{E}_z \left[ \log (D(G(\mathbf{0}_z | x))) \right] + \lambda \lVert G(\mathbf{0}_z | x) - y \rVert_1
	\end{equation}
	\begin{equation}
		\max_{\theta_D} J^{(D)} (\theta_D, \theta_G) = \max_{\theta_D} \left( \mathbb{E}_y \left[ \log (D(y|x)) \right] + \mathbb{E}_z \left[ \log (1 - D(G(\mathbf{0}_z | x) | x)) \right] \right)
	\end{equation}
	The discriminator gets colored images from both generator and original data along with the grayscale input as the condition and tries to decide which pair contains the true colored image.

\section{Method}

	Image colorization is an image-to-image translation problem that maps a high dimensional input to a high dimensional output. It can be seen as pixel-wise regression problem where structure in the input is highly aligned with structure in the output. That means the network needs not only to generate an output with the same spatial dimension as the input, but also to provide color information to each pixel in the grayscale input image. We provide an entirely convolutional model architecture using a regression loss as our baseline and then extend the idea to adversarial nets. \\
	In this work we utilize the L*a*b* color space for the colorization task. This is because L*a*b* color space contains dedicated channel to depict the brightness of the image and the color information is fully encoded in the remaining two channels. As a result, this prevents any sudden variations in both color and brightness through small perturbations in intensity values that are experienced through RGB.

\subsection{Baseline Network}

\begin{figure}[!htb]
		\centering
		\includegraphics[width=1\textwidth]{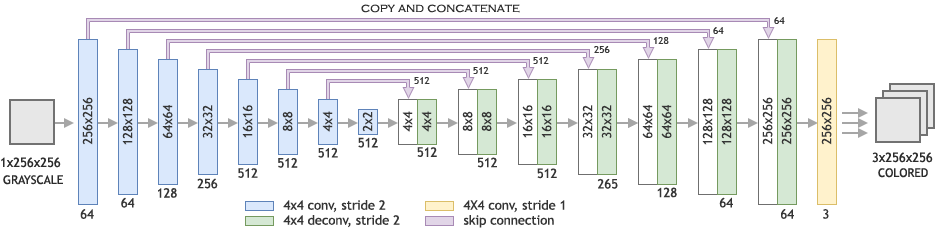}
		\caption{U-Net architecture ($256 \times 256$ input) }
		\label{fig:unet}
	\end{figure}

	For our baseline model, we follow the ``fully convolutional network"\cite{long2015fully} model where the fully connected layers are replaced by convolutional layers which include upsampling instead of pooling operators. This idea is based on encoder-decoder networks \cite{hinton2006reducing} where input is progressively downsampled using a series of contractive encoding layers, and then the process is reversed using a series of expansive decoding layers to reconstruct the input. Using this method we can train the model end-to-end without consuming large amounts of memory. Note that the subsequent downsampling leads to a much more compact feature learning in the middle layers. This strategy forms a crucial attribute to the network, otherwise the resolution would be limited by GPU memory. \\
	Our baseline model needs to find a direct mapping from the grayscale image space to color image space. However, there is an information bottleneck that prevents flow of the low level information in the network in the encoder-decoder architecture. To fix this problem, features from the contracting path are concatenated with the upsampled output in the expansive path within the network. This also makes the input and output share the locations of prominent edges in grayscale and colored images. This architecture is called U-Net \cite{ronneberger2015u}, where skip connections are added between layer \textit{i} and layer \textit{n-i}. \\
	The architecture of the model is symmetric, with $n$ encoding units and $n$ decoding units. The contracting path consists of $4 \times 4$ convolution layers with stride 2 for downsampling, each followed by batch normalization \cite{ioffe2015batch} and Leaky-ReLU \cite{maas2013rectifier} activation function with the slope of 0.2. The number of channels are doubled after each step. Each unit in the expansive path consists of a $4 \times 4$ transposed convolutional layer with stride 2 for upsampling, concatenation with the activation map of the mirroring layer in the contracting path, followed by batch normalization and ReLU activation function. The last layer of the network is a $1 \times 1$ convolution which is equivalent to cross-channel parametric pooling layer. We use $\tanh$ function for the last layer as proposed by \cite{isola2016image}. The number of channels in the output layer is 3 with L*a*b* color space. (Fig. \ref{fig:unet})
	\vspace{-10px}
	\vspace{-20px}
	\\We train the baseline model to minimize the Euclidean distance between predicted and ground truth averaged over all pixels: 
	\begin{equation}
		J(x;\theta) = \frac{1}{3n} \sum_{\ell = 1}^3 \sum_{p=1}^n \Vert h(x;\theta)^{(p,\ell)} - y^{(p,\ell)} \Vert_2^2
	\end{equation}
	where $x$ is our grayscale input image, $y$ is the corresponding color image, $p$ and $\ell$ are indices of pixels and color channels respectively, $n$ is the total number of pixels, and $h$ is a function mapping from grayscale to color images.

\subsection{Convolutional GAN}

	For the generator and discriminator models, we followed Deep Convolutional GANs (DCGAN) \cite{radford2015unsupervised} guidelines and employed convolutional networks in both generator and discriminator architectures. The architecture was also modified as a conditional GAN instead of a traditional DCGAN; we also follow guideline in \cite{isola2016image} and provide noise only in the form of dropout \cite{srivastava2014dropout}, applied on several layers of our generator. The architecture of generator $G$ is the same as the baseline. For discriminator $D$, we use similar architecture as the baselines contractive path: a series of $4 \times 4$ convolutional layers with stride 2 with the number of channels being doubled after each downsampling. All convolution layers are followed by batch normalization, leaky ReLU activation with slope 0.2. After the last layer, a convolution is applied to map to a 1 dimensional output, followed by a sigmoid function to return a probability value of the input being real or fake. The input of the discriminator is a colored image either coming from the generator or true labels, concatenated with the grayscale image. 

\subsection{Training Strategies}

	For training our network, we used Adam \cite{kingma2014adam} optimization and weight initialization as proposed by \cite{he2015delving}. We used initial learning rate of $2 \times 10^{-4}$ for both generator and discriminator and manually decayed the learning rate by a factor of 10 whenever the loss function started to plateau. For the hyper-parameter $\lambda$ we followed the protocol from \cite{isola2016image} and chose $\lambda = 100$, which forces the generator to produce images similar to ground truth.\\
	GANs have been known to be very difficult to train as it requires finding a Nash equilibrium of a non-convex game with continuous, high dimensional parameters \cite{salimans2016improved}. We followed a set of constraints and techniques proposed by \cite{isola2016image,radford2015unsupervised,salimans2016improved,creswell2017generative} to encourage convergence of our convolutional GAN and make it stable to train.
	\begin{itemize}
		\item \textbf{Alternative Cost Function} \\
		This heuristic alternative cost function \cite{goodfellow2016nips} was selected due to its non-saturating nature; the motivation for this cost function is to ensure that each player has a strong gradient when that player is ``losing" the game.
		
		\item \textbf{One Sided Label Smoothing} \\
		Deep neural networks normally tend to produce extremely confident outputs when used in classification. It is shown that replacing the 0 and 1 targets for a classifier with smoothed values, like .1 and .9 is an excellent regularizer for convolutional networks \cite{szegedy2016rethinking}. Salimans et al \cite{salimans2016improved} demonstrated that \textit{one-sided label smoothing} will encourage the discriminator to estimate soft probabilities and reduce the vulnerability of GANs to adversarial examples. In this technique we smooth \textit{only} the positive labels to 0.9, leaving negative labels set to 0.
		
		\item \textbf{Batch Normalization} \\
		One of the main difficulties when training GANs is for the generator to collapse to a parameter setting where it always emits the same output \cite{salimans2016improved}. This phenomenon is called \textit{mode-collapse}, also known as \textbf{the Helvetica scenario} \cite{goodfellow2016nips}. When mode-collapse has occurred, the generator learns that a single output is able to consistently trick the discriminator. This is non-ideal as the goal is for the network to learn the distribution of the data rather than the most ideal way of fooling the discriminator. Batch normalization \cite{ioffe2015batch} is proven to be essential to train both networks preventing the generator from collapsing all samples to a single point \cite{radford2015unsupervised}. Batch-Norm is not applied on the first layer of generator and discriminator and the last layer of the generator as suggested by \cite{isola2016image}.
		
		\item \textbf{All Convolutional Net} \\
		Strided convolutions are used instead of spatial pooling functions. This effectively allows the model to learn its own downsampling/upsampling rather than relying on a fixed downsampling/upsampling method. This idea was proposed in \cite{springenberg2014striving} and has shown to improve training performance as the network learns all necessary invariances just with convolutional layers.
	
		\item \textbf{Reduced Momentum} \\
		We use Adam optimizer \cite{kingma2014adam} for training both networks. Recent research has shown that using a large momentum term $\beta_1$ (0.9 as suggested), could result in oscillation and instability in training. We followed the suggestion in \cite{radford2015unsupervised} to reduce the momentum term to 0.5.
		
		\item \textbf{LeakyReLU Activation Function}  \\
		Radford et al. \cite{radford2015unsupervised} showed that using leaky ReLU \cite{isola2016image} activation functions in the discriminator resulted in better performance over using regular ReLUs. We also found that using leaky ReLU in the encoder part of the generator as suggested by \cite{isola2016image} works slightly better.
	\end{itemize}

\section{Experimental Results}

	To measure the performance, we have chosen to employ mean absolute error (MAE) and accuracy. MAE is computed by taking the mean of the absolute error of the generated and source images on a pixel level for each color channel. Accuracy is measured by the ratio between the number of pixels that have the same color information as the source and the total number of pixels. Any two pixels are considered to have the same color if their underlying color channels lie within some threshold distance $\epsilon$. This is mathematically represented by
	\begin{equation}
		acc(x,y) = \frac{1}{n} \sum_{p = 1}^n \prod_{\ell=1}^3 \mathbf{1}_{[0,\epsilon_{\ell}]}(\vert h(x)^{(p, \ell)} - y^{(p, \ell)} \vert)
	\end{equation}
	where $\mathbf{1}_{[0,\epsilon_{\ell}]}(\mathbf{x}), \mathbf{x} \in \mathbb{R}$ denotes the indicator function, $y$ is the corresponding color image, $h$ is a function mapping from grayscale to color images, and $\epsilon_{\ell}$ is a threshold distance used for each color channel. The training results for each model are summarized in Table \ref{tab:results}.
	\begin{table}
		\def\arraystretch{1.2}
		\centering
		\begin{tabular}{|c|c|c|c|c|c|c|c|}
			\hline
			\textbf{Dataset} & \textbf{Network} & \textbf{Batch Size} & \textbf{EPOCHs} & \textbf{MAE} & \textbf{Accuracy $\epsilon=2\%$} & \textbf{Accuracy $\epsilon=5\%$} \\ \hline
			CIFAR-10 & U-Net & 128 & 200 & 7.9 & 13.7 & 37.2\% \\ \hline
			CIFAR-10 & GAN & 128 & 200 & 5.1 & 24.1 & 65.5\% \\ \hline
			Places365 & GAN & 16 & 20 & 7.5 & 18.3 & 47.3 \% \\ \hline
		\end{tabular}
		\vspace{5px}
		\caption{Training results of baseline model and GAN. }
		\label{tab:results}
	\end{table}
	Some of the preliminary results using the CIFAR-10 ($32 \times 32$) dataset are shown in Appendix A. The images from GAN had a clear visual improvement than those generated by the baseline CNN. The images generated by GAN contained colors that were more vibrant whereas the results from CNN suffered from a light hue. In some cases, the GAN was able to nearly replicate the ground truth. However, one drawback was that the GAN tends to colorize objects in colors that are most frequently seen. For example, many car images were colored red. This is most likely due to the significantly larger number of images with red cars than images with cars of another color. \\
	The preliminary results using Places365 ($256 \times 256$) are shown in Appendix B. We noticed that there were some instances of mis-colorization: regions of images that have high fluctuations are frequently colored green. This is likely caused by the large number of grassland images in the training set, thus the model leans towards green whenever it detects a region with high fluctuations in pixel intensity values. We also noticed that some colorized images experienced a ``sepia effect'' seen with CIFAR-10 under U-Net. This hue is evident especially with images with clear sky, where the color of the sky includes a strange color gradient between blue and light yellow. We suspect that this was caused by insufficient training, and will correct itself over time.

\section{Conclusion and Future Work}

	In this study, we were able to automatically colorize grayscale images using GAN, to an acceptable visual degree. With the CIFAR-10 dataset, the model was able to consistently produce better looking (qualitatively) images than U-Net. Many of the images generated by U-Net had a brown-ish hue in the results known as the ``Sepia effect'' across L*a*b* color space. This is due to the L2 loss function that was applied to the baseline CNN, which is known to cause a blurring effect. \\
	We obtained mixed results when colorizing grayscale images using the Places365 dataset. Mis-colorization was a frequent occurrence with images containing high levels of textured details. This leads us to believe that the model has identified these regions as grass since many images in the training set contained leaves or grass in an open field. In addition, this network was not as well-trained as the CIFAR-10 counterpart due to its significant increase in resolution ($256 \times 256$ versus $32 \times 32$) and the size of the dataset (1.8 million versus $50,000$). We expect the results will improve if the network is trained further. \\
	We would also need to seek a better quantitative metric to measure performance. This is because all evaluations of image quality were qualitative in our tests. Thus, having a new or existing quantitative metric such as peak signal-to-noise ratio (PSNR) and root mean square error (RMSE) will enable a much more robust process of quantifying performance. \\
	\noindent
	Source code is publicly available at:
	\vskip 2mm
	\url{https://github.com/ImagingLab/Colorizing-with-GANs}

\small{\subsubsection*{Acknowledgments}
{This research was supported in part by an NSERC Discovery Grant for ME. The authors gratefully acknowledge the support of NVIDIA Corporation for donation of GPUs through its Academic Grant Program.}}

\vspace{-5px}
\addcontentsline{toc}{section}{Bibliography}
\bibliography{colorization_gan}{}
\bibliographystyle{unsrt}

\newpage
\appendix
\section{CIFAR-10 Results}
\vspace{-20px}

	\begin{figure}[!htb]
		\centering
		\includegraphics[width=8.6cm]{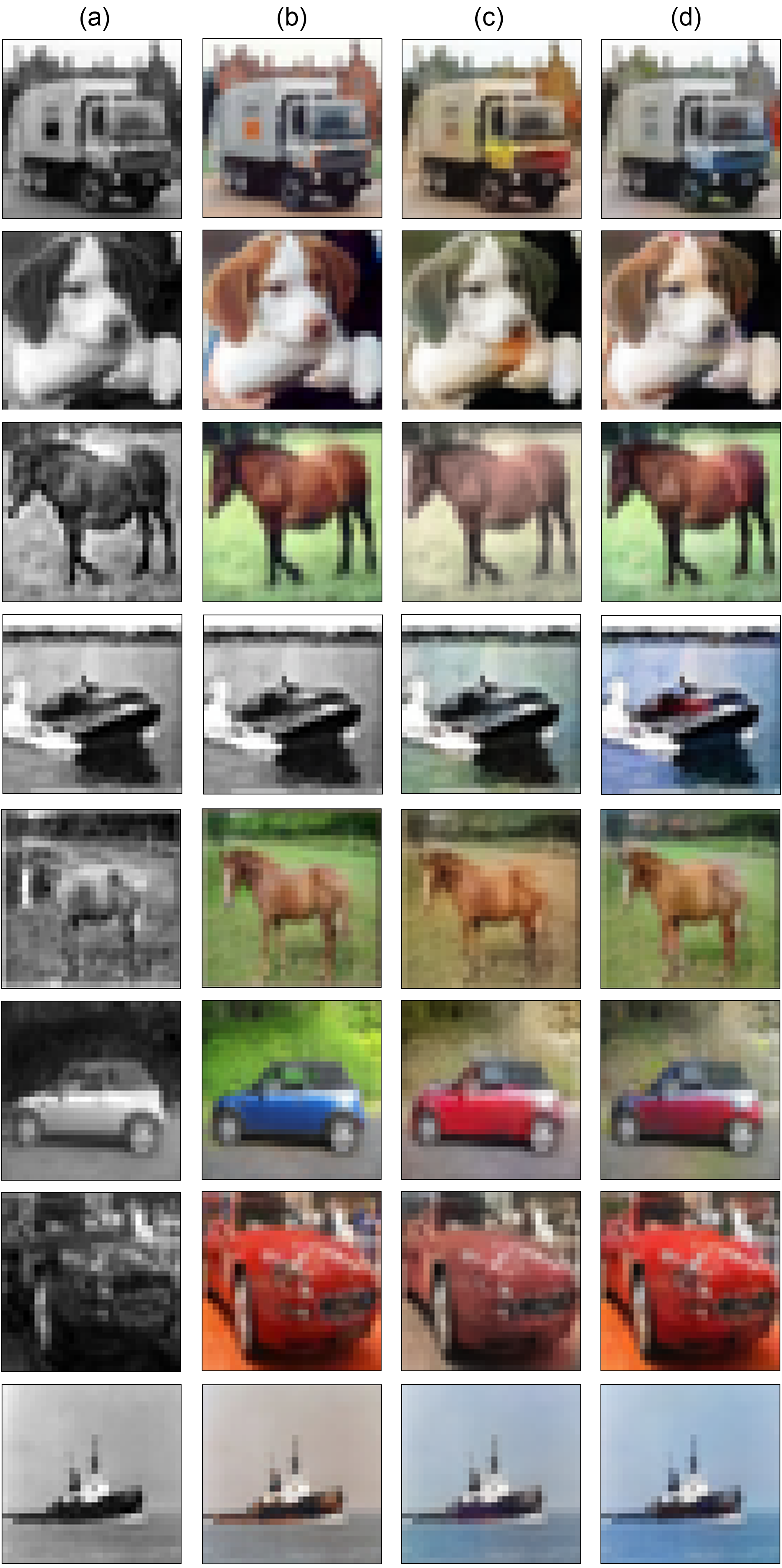}
		\caption{Colorization results with CIFAR10. (a) Grayscale. (b) Original Image. (c) Colorized with U-Net. (d) Colorized with GAN.}
	\end{figure}

\newpage
\section{Places365 Results}
\vspace{-20px}

	\begin{figure}[!htb]
		\centering
		\includegraphics[width=12cm]{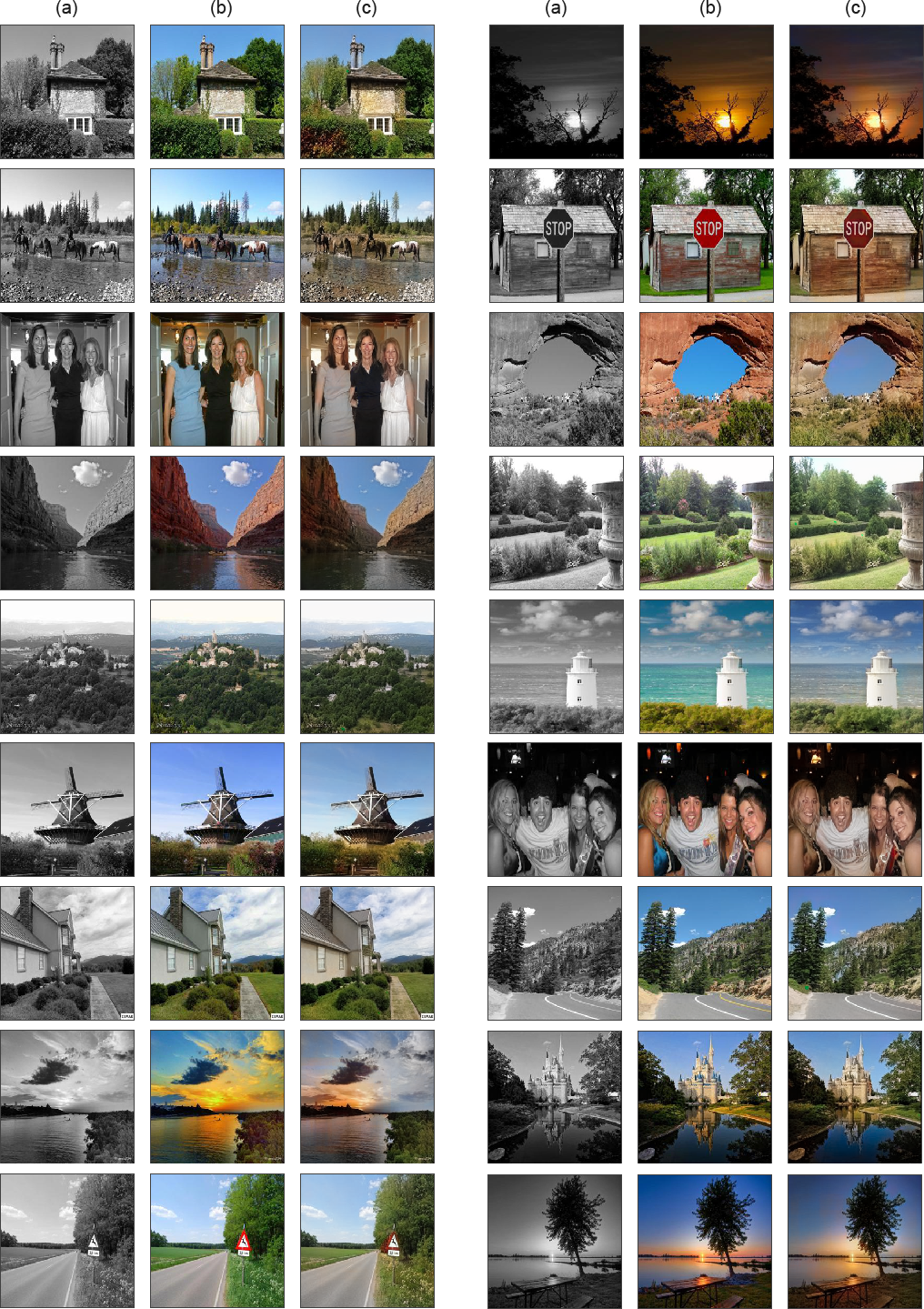}
		\caption{Colorization results with Places365 (a) Grayscale. (b) Original Image. (c) Colorized with GAN.}
	\end{figure}

\end{document}